%% file: main.tex
\documentclass[10pt,twocolumn,letterpaper]{article}

\usepackage[pagenumbers]{cvpr} 

\input{preamble}

\definecolor{cvprblue}{rgb}{0.21,0.49,0.74}
\usepackage[pagebackref,breaklinks,colorlinks,citecolor=cvprblue]{hyperref}

\def\paperID{*****} 
\def\confName{CVPR}
\def\confYear{2024}

\title{EdgeVidSum: Real-Time Personalized Video Summarization at the Edge}

\author{Ghulam Mujtaba\\
Anderson College of Business and Computing\\
Regis University, Denver, CO, USA\\
{\tt\small gmujtaba@ieee.org}
\and
Eun-Seok Ryu\\
Department of Immersive Media Engineering \\
Sungkyunkwan University, Seoul, Korea\\
{\tt\small esryu@skku.edu}
}

\begin{document}
\maketitle
\input{sec/0_abstract}    
\input{sec/1_intro}

{
    \small
    \bibliographystyle{ieeenat_fullname}
    \bibliography{main}
}


\end{document}

%% file: preamble.tex
\usepackage[dvipsnames]{xcolor}


%% file: sec/0_abstract.tex
\begin{abstract}
EdgeVidSum is a lightweight method that generates personalized, fast-forward summaries of long-form videos directly on edge devices. The proposed approach enables real-time video summarization while safeguarding user privacy through local data processing using innovative thumbnail-based techniques and efficient neural architectures. Unlike conventional methods that process entire videos frame by frame, the proposed method uses thumbnail containers to significantly reduce computational complexity without sacrificing semantic relevance. The framework employs a hierarchical analysis approach, where a lightweight 2D CNN model identifies user-preferred content from thumbnails and generates timestamps to create fast-forward summaries. Our interactive demo highlights the system's ability to create tailored video summaries for long-form videos, such as movies, sports events, and TV shows, based on individual user preferences. The entire computation occurs seamlessly on resource-constrained devices like Jetson Nano, demonstrating how EdgeVidSum addresses the critical challenges of computational efficiency, personalization, and privacy in modern video consumption environments.
\end{abstract}

%% file: sec/1_intro.tex
\section{Introduction}\label{sec:intro}
Over-the-top (OTT) streaming services, such as Netflix and Amazon Prime, have transformed digital media consumption by providing extensive libraries of long-form content \cite{vernon2024future}. These platforms have revolutionized how users interact with video content across multiple devices and viewing contexts. While high-speed internet and smart devices have increased accessibility, efficiently navigating lengthy (long-form) videos remains a challenging task. The abundance of content has created information overload for many viewers. Users often seek ways to capture key moments of interest without watching the entire content. This need is particularly acute for educational videos, sports broadcasts, and documentary content, where specific segments are more relevant than others.

\begin{figure}[t]
\centering
\includegraphics[width=\linewidth]{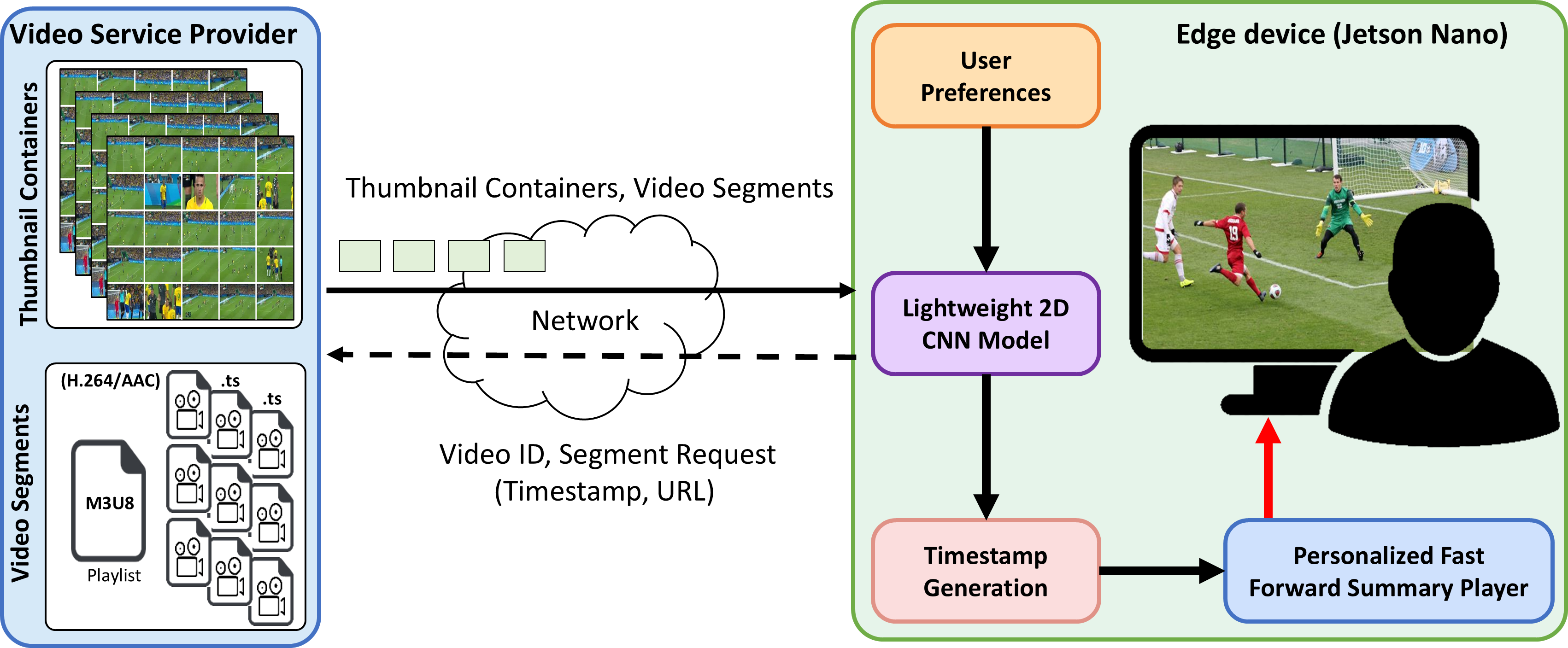}
\caption{System architecture of EdgeVidSum showing the flow from thumbnail container generation to personalized summary playback.}
\label{fig:system}
\end{figure}

Automatic video summarization has emerged as a solution to reduce playback time while preserving critical information from the original video \cite{mujtaba2022ltc}. This method enables viewers to consume essential content in significantly reduced timeframes while maintaining narrative coherence. Traditional video summarization approaches can be categorized into keyframe-based \cite{huang2019novel} and keyshot-based methods \cite{song2016click}. Keyframe methods identify informative frames to create representative image sets through techniques such as clustering algorithms, attention mechanisms, and importance scoring \cite{huang2019novel}. Keyshot methods segment videos into short clips to preserve motion dynamics through boundary detection and semantic segmentation approaches \cite{song2016click}. These techniques better maintain action continuity but require sophisticated algorithms to identify coherent segments.

However, these methods often introduce temporal discontinuities, making it difficult for viewers to follow the context between segments. The disconnected nature of these summaries can result in fragmented narratives that fail to convey the complete story. For example, the viewer may observe a player celebrating a goal in a football match without understanding the important sequence of actions that led to the scoring opportunity. Fast-forward summarization (FF-SUM) \cite{mujtaba2025personalized} addresses this issue by accelerating video playback while maintaining temporal continuity. FF-SUM uses variable-speed playback techniques that dynamically adjust frame rates based on the importance of the content. This approach ensures viewers experience continuous playback without jarring transitions while spending less time on less relevant content.

Existing methods produce generic video summaries without considering individual preferences \cite{mujtaba2023frc}. These approaches generate identical summaries for all viewers despite their diverse interests. For example, during a soccer match, one user may prefer to watch goal highlights, while another focuses on defensive plays. Personalized FF-SUM addresses this limitation by dynamically selecting semantically significant segments based on individual preferences \cite{mujtaba2023frc, mujtaba2025personalized}. This approach integrates user profiles, explicit preference settings, and implicit feedback mechanisms to understand viewer interests. The system analyzes viewing patterns and engagement metrics to identify segments matching specific user preferences. 

Implementing personalized FF-SUM on edge devices poses significant challenges in terms of both privacy and computational efficiency. Traditional approaches often depend on cloud-based processing, which involves transmitting user data to external servers \cite{ali2015security}. Server-side solutions can reduce the workload on edge devices, but they raise significant privacy concerns and may lead to inconsistent performance due to changing network conditions. Additionally, user preferences and viewing habits often contain sensitive information that people might not want to share. This makes on-device processing a more essential solution.

This paper introduces EdgeVidSum, a novel client-driven, personalized FF-SUM method that efficiently summarizes long-form videos on resource-constrained devices, such as the Jetson Nano. The system uses lightweight neural networks optimized for edge deployment. Figure \ref{fig:system} shows the proposed system architecture of the EdgeVidSum method. The proposed approach minimizes computational overhead by using lightweight thumbnail containers instead of processing entire videos, allowing for real-time summarization while maintaining semantic relevance. These thumbnail containers serve as compressed video representations that preserve essential visual and temporal information while requiring significantly less storage and processing power. The proposed method enhances privacy and reduces latency by conducting the entire process locally on the device.

\section{Related Work}

Video summarization has evolved from traditional content-based approaches to more sophisticated techniques. Early methods employed visual attention modeling, clustering, and temporal segmentation to identify salient content \cite{ngo2005video, ma2005generic, song2015tvsum}. Graph-based models emerged to detect and group meaningful events \cite{peng2006clip}, while supervised learning techniques introduced category-specific summaries using trained classifiers \cite{potapov2014category}. Multi-objective optimization frameworks balanced relevance, diversity, and coverage \cite{gygli2015video}, and gaze information enhanced egocentric video summarization by leveraging fixation cues \cite{xu2015gaze}. Recent personalized approaches have significantly advanced the field by incorporating user preferences. Methods like \cite{varini2017personalized} integrated geolocation data with semantic knowledge for customized egocentric video summaries, while \cite{sharghi2017query} introduced query-focused summarization using memory networks. Vision-language embeddings enabled freeform text input for customization \cite{plummer2017enhancing}, and multimodal information from image collections helped mine personal interests \cite{yin2016encoded}. Reinforcement learning frameworks offered scalable solutions by optimizing user-defined preferences such as faces and scene diversity \cite{nagar2021generating}.

Fast-forward video summarization focuses on accelerating playback while maintaining semantic integrity and visual coherence. Several techniques employ reinforcement learning for frame selection, such as FFNet \cite{lan2018ffnet}, which formulates the task as a Markov decision process using Q-learning. Sparse Adaptive Sampling (SAS) \cite{silva2020sparse} reduces visual discontinuities by inserting frames during abrupt scene transitions. Multimodal approaches like Musical Hyperlapse \cite{de2023multimodal} align playback speed with background music rhythm, creating emotionally synchronized summaries. User-centric methods leverage social network data to infer interests for personalized fast-forward videos \cite{ramos2020personalizing}, while neural network-based approaches like SpeedNet \cite{benaim2020speednet} predict object ``speediness'' for adaptive fast-forwarding with minimal artifacts. Despite these advancements, significant challenges remain. Many current techniques struggle to balance personalization with temporal consistency, resulting in fragmented summaries. Fast-forward methods often introduce visual artifacts or jerky transitions \cite{silva2020sparse}, and most approaches require extensive computational resources unsuitable for edge devices. Although some work \cite{wei2007video, xu2008personalized} has explored video personalization for resource-constrained environments, there remains a need for efficient, edge-compatible frameworks that generate personalized and temporally coherent fast-forward summaries while preserving narrative flow.

\section{Method}

\subsection{Thumbnail-Based Approach}
Processing entire videos frame by frame for summarization is computationally expensive, especially on resource-constrained devices. For a one-minute video at 25 FPS, 1500 frames must be analyzed, with significant redundancy between consecutive frames. The proposed approach uses lightweight thumbnail containers instead of full-resolution frames to reduce computational requirements dramatically. This approach achieves a 97.8\% reduction in data processing volume compared to traditional frame-by-frame methods.

Thumbnails offer several advantages compared to video frames. First, they maintain a consistent lightweight size (160×90 pixels) regardless of original video resolution, reducing memory requirements by approximately 98.2\% compared to full HD frames. Second, thumbnails exist in significantly reduced quantities. For example, a one-hour 52-minute football video has approximately \emph{202,036} frames but only \emph{6,734} thumbnails, improving processing speed by a factor of 30. Third, thumbnails provide higher quality representation through intelligent selection algorithms that maintain semantic integrity \cite{mujtaba2020client}. Finally, they use an optimized storage format that requires only 4 KB per thumbnail, compared to 1-3 MB for full-resolution frames.

\begin{figure*}[t]
\centering
\includegraphics[width=\linewidth]{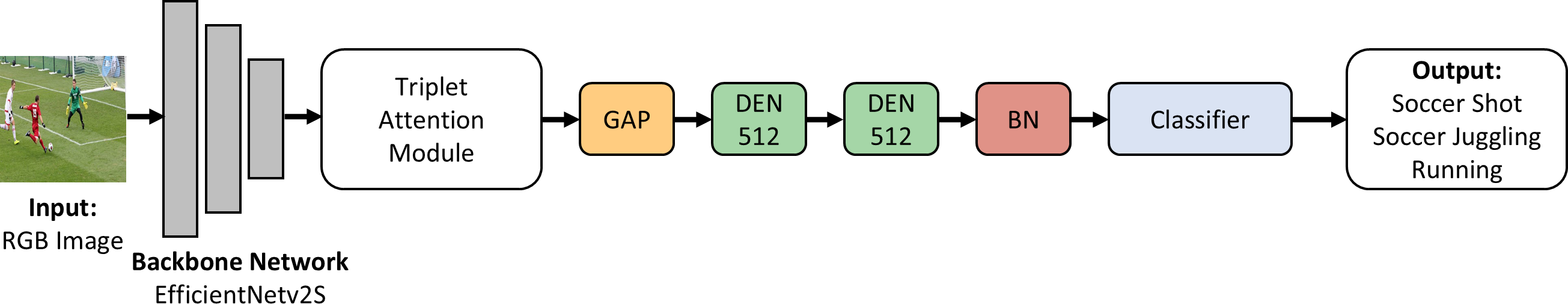}
\caption{Complete architecture of our Thumbnail Container Analyzer. The RGB input image is processed by the EfficientNetV2-S backbone network, followed by our Triplet Attention Module. The attention-enhanced features undergo Global Average Pooling (GAP), followed by two dense layers (DEN) of 512 units each with ReLU activation and 0.2 dropout. A Batch Normalization (BN) layer precedes the final classifier with softmax activation, which outputs category predictions (e.g., Soccer Shot, Soccer Juggling, Running).}
\label{fig:network}
\end{figure*}

\begin{figure}[t]
\centering
\includegraphics[width=\linewidth]{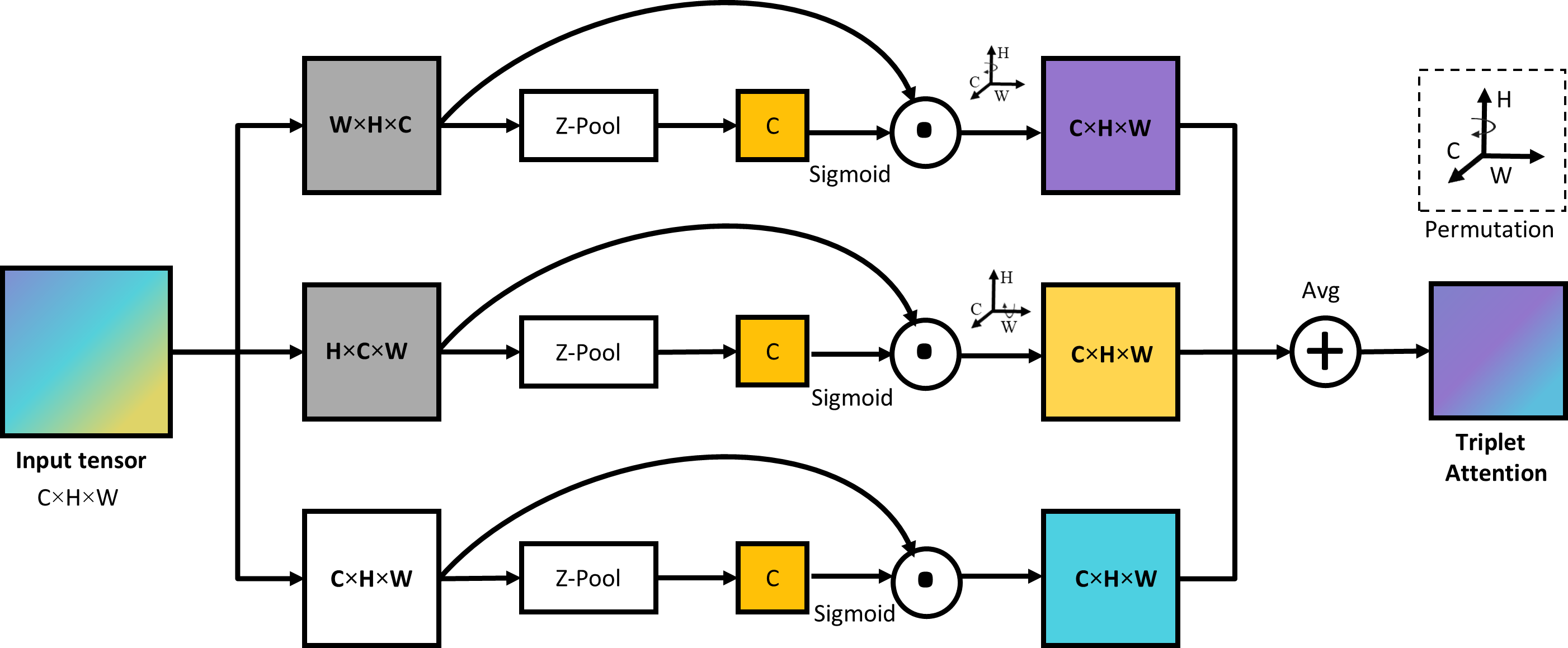}
\caption{Detailed architecture of the Triplet Attention Module (TAM). The input tensor undergoes three parallel processing branches that capture different cross-dimensional interactions: width-channel (top), height-channel (middle), and height-width (bottom). Each branch applies Z-pooling (combining max and average pooling), convolution with sigmoid activation, and element-wise multiplication (denoted by $\odot$). The outputs from all three branches are averaged to produce the final attention-refined feature representation.}
\label{fig:triplet}
\end{figure}

\subsection{Thumbnail Container Analyzer}\label{subsec:tca_section}

Our Thumbnail Container Analyzer architecture comprises three primary components: a feature extraction backbone, an attention mechanism, and a classification head (Fig. \ref{fig:network}). For the backbone, we employ EfficientNetV2 base \cite{tan2104efficientnetv2} pre-trained on ImageNet \cite{deng2009imagenet}. This model was selected due to its favorable accuracy-to-computation ratio when extracting high-level semantic representations from thumbnail images. The extracted features subsequently flow into our attention module, which enhances contextual information capture. The final classification stage processes these attention-refined features for prediction.

\subsubsection{Triplet Attention Module}
We implement a Triplet Attention Module (TAM) \cite{TAM} before the global average pooling layer in our architecture. TAM architecture is illustrated in Fig. \ref{fig:triplet}. The module functions through three parallel branches that capture cross-dimensional interactions across the feature space. The first branch models interactions between width and channel dimensions. The second branch focuses on height and channel dimensional relationships. The third branch addresses width and height interactions, effectively implementing spatial attention.

The operational mechanics of TAM can be formalized mathematically. We denote the input feature map from the EfficientNetV2 backbone as $\Psi_{i} \in \mathbf{R}^{C \times H \times W}$, where $C$, $H$, and $W$ represent channel count, height, and width dimensions, respectively. For the first branch capturing channel-width interactions, we perform:

\begin{equation}
\Psi_{b1} = \text{concat}(\text{pool}{max}^{H}(\Psi{i}), \text{pool}{avg}^{H}(\Psi{i}))
\end{equation}

This operation applies maximum and average pooling along the height dimension, concatenating the results to produce $\Psi_{b1}$ with dimensions $\mathbf{R}^{2 \times C \times W}$. We then process this representation through a convolutional layer with $7 \times 7$ kernels followed by sigmoid activation:

\begin{equation}
\mathcal{F}{b1} = \sigma(\rho(\Psi{b1}))
\end{equation}

Here, $\sigma$ represents the sigmoid function and $\rho$ denotes the convolution operation. Analogous processes generate $\mathcal{F}{b2}$ and $\mathcal{F}{b3}$ for the width-channel and height-width interactions respectively. The final TAM output combines these attention maps:

\begin{equation}
\Psi_{TAM} = \frac{(\Psi_{i} \odot \mathcal{F}{b1} + \Psi{i} \odot \mathcal{F}{b2} + \Psi{i} \odot \mathcal{F}_{b3})}{3}
\end{equation}

The symbol $\odot$ indicates element-wise multiplication. Our implementation of TAM was motivated by its efficient integration of channel and spatial attention while preserving spatial information integrity. This approach aligns with our research focus on lightweight personalized video summarization. TAM offers computational efficiency with only 0.3M additional parameters while delivering a 4.2\% classification accuracy improvement in our experimental evaluation, compared to alternatives such as channel attention \cite{SEN} and convolutional block attention modules \cite{SEN}.

\subsubsection{Classification Head}
Following the TAM, we employ a Global Average Pooling (GAP) layer to reduce spatial dimensions while preserving channel information. The pooled features are processed through two dense layers, each containing 512 units with ReLU activation and a dropout rate of 0.2. These layers enhance feature abstraction and mitigate overfitting. A Batch Normalization layer is inserted before the final classifier to stabilize training. The classification head uses softmax activation to produce probability distributions across the target activity categories. This entire pipeline is optimized end-to-end using cross-entropy loss.

\subsection{System Architecture}
EdgeVidSum employs a fully client-driven approach that leverages the computational resources of edge devices while protecting user privacy. As illustrated in Figure \ref{fig:system}, the architecture consists of two main components: the Video Service Provider (VSP) and the Edge Device (Jetson Nano). The system operates as follows:

\begin{enumerate}
    \item \textbf{Thumbnail Container Generation}: The client application requests summarization for a selected video. The VSP hosts two key repositories: a thumbnail container database that stores visual representations of videos, and a video segments database containing the actual video content in various formats, including HLS, DASH, AAC, and MP4. When a request is received, the VSP transmits pre-generated thumbnail containers representing the entire video at a very low bitrate \cite{mujtaba2020client}. These containers are efficiently transferred over the network to the edge device.
    
    \item \textbf{Preference-Based Selection}: User preferences can be explicitly specified through the interface or implicitly inferred from viewing history. Thumbnails will be processed using a lightweight 2D CNN model on an edge device based on these selections. Details of the model are provided in Section \ref{subsec:tca_section}. This model processes the thumbnail containers to identify those that align with the user's preferences. It efficiently detects specific scenes, events, or objects of interest within the content while operating within the computational limits of the Jetson Nano.
        
    \item \textbf{Timestamp Generation}: Once preferred content is identified, the system generates precise timestamps corresponding to segments of interest. The client narrows temporal periods of interest by analyzing thumbnail containers at increasing temporal granularity until reaching segment-level precision. These timestamps define the boundaries of content segments that align with user preferences.
    
    \item \textbf{Personalized Fast Forward Summary Player}: Using the generated timestamps, the client sends requests to the VSP for specific video segments identified by their Video ID, Timestamp, and URL. The personalized player on the edge device then manages the playback speed based on content relevance, playing preferred segments at normal speed while accelerating through less critical content. This adaptive playback creates a seamless viewing experience tailored to individual preferences, displayed on the edge device's screen.
\end{enumerate}

This architecture maintains data privacy by performing all preference analysis and summarization decisions locally on the edge device. Only essential communication, such as thumbnail containers and specific video segment requests, occurs between the client and the VSP, significantly reducing network bandwidth requirements while providing a personalized viewing experience.

\section{Implementation and Demo}
Our interactive demo showcases EdgeVidSum running on a Jetson Nano device (Quad-core ARM A57 CPU, 128-core Maxwell GPU, 4GB RAM), demonstrating real-time personalized summarization across various long-form video genres, including sports, movies, and documentaries. The demonstration highlights several key features and capabilities. The system allows users to select their preferences through a simple interface with both categorical and fine-grained options. It performs real-time thumbnail analysis using our lightweight 2D CNN model. EdgeVidSum dynamically generates personalized fast-forward summaries with variable playback rates. The system delivers seamless playback with adaptive speed control using a custom-developed video rendering pipeline. Privacy is preserved through complete on-device processing with any private data transmission. 

\begin{figure}[t]
\centering
\includegraphics[width=\linewidth]{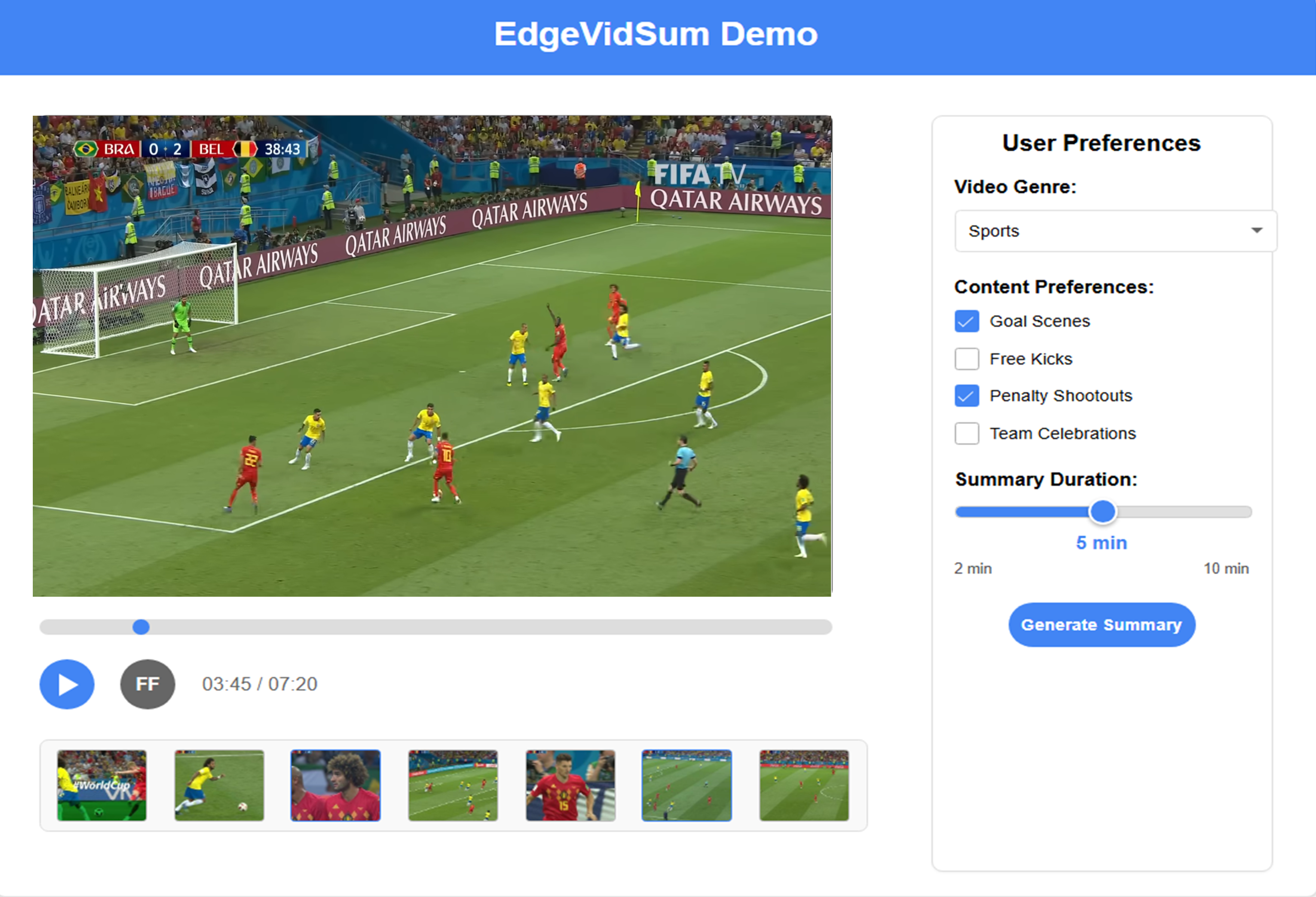}
\caption{Screenshot of EdgeVidSum demo interface showing preference selection and personalized summary playback.}
\label{fig:demo}
\end{figure}

The lightweight model accurately identifies user-preferred content while maintaining a minimal computational footprint. Our implementation achieves summarization speeds suitable for real-time interaction, with processing times ranging from 2 to 5 seconds for 90-minute videos, depending on the content complexity. Figure \ref{fig:demo} shows the demo interface, showing preference selection and personalized summary playback.

\section{Conclusion}

EdgeVidSum presents a novel approach to personalized video summarization that addresses key limitations of existing methods. By leveraging lightweight thumbnail containers and efficient neural architectures, our framework enables real-time summarization directly on edge devices without compromising user privacy. This approach is particularly valuable in today's streaming ecosystem, where users demand personalized experiences without sacrificing efficiency or privacy. Our demonstration showcases the practical application of EdgeVidSum across various content types, highlighting its ability to deliver tailored summaries that align with individual preferences while operating within the constraints of edge computing environments. Future work will focus on expanding the range of detectable preferences and optimizing the neural architecture for even more efficient operation.

\section{Acknowledgements}
This research was supported by the Ministry of Science, ICT (MSIT), Korea, under the Global Scholars Invitation Program (RS-2025-00459638) supervised by the Institute for Information \& Communications Technology Planning \& Evaluation (IITP). It was also supported by the Ministry of Science and ICT (MSIT), Korea, under the Graduate School of Metaverse Convergence support program (IITP-2025-RS-2023-00254129) supervised by the Institute for Information \& Communications Technology Planning \& Evaluation (IITP).